\documentclass{article}

\usepackage{arxiv}
\usepackage[numbers]{natbib}

\usepackage[utf8]{inputenc} 
\usepackage[T1]{fontenc}    
\usepackage{hyperref}       
\usepackage{url}            
\usepackage{booktabs}       
\usepackage{amsfonts}       
\usepackage{nicefrac}       
\usepackage{microtype}      
\usepackage{lipsum}		
\usepackage{graphicx}
\usepackage{natbib}
\usepackage{doi}
\usepackage{longtable}
\usepackage{array}

\title{Invoice Information Extraction: Methods and Performance Evaluation}

\renewcommand{\headeright}{}
\renewcommand{\undertitle}{}

\author{
Sai Yashwant \\
Technology Product Leader – Data Science and AI \\
ManpowerGroup Services India Pvt. Ltd. \\
\texttt{sai.yashwant@manpowergroup.com}
\And
Anurag Dubey \\
Executive Data Scientist \\
ManpowerGroup Services India Pvt. Ltd. \\
\texttt{anurag.dubey@manpowergroup.com}
\And
Praneeth Paikray \\
Senior Gen AI Specialist \\
ManpowerGroup Services India Pvt. Ltd. \\
\texttt{praneeth.paikray@manpowergroup.com}
\And
Gantala Thulsiram \\
Assistant Professor \\
Indian Institute of Technology, Hyderabad \\
\texttt{thulsiramg@mae.iith.ac.in}
}
   
\hypersetup{
pdftitle={A template for the arxiv style},
pdfsubject={q-bio.NC, q-bio.QM},
pdfauthor={David S.~Hippocampus, Elias D.~Striatum},
pdfkeywords={First keyword, Second keyword, More},
}

\begin{document}
\maketitle
\renewcommand{\headeright}{}
\renewcommand{\undertitle}{}

\begin{abstract}
This paper presents methods for extracting structured information from invoice documents and proposes a set of evaluation metrics (EM) to assess the accuracy of the extracted data against annotated ground truth. The approach involves pre-processing scanned or digital invoices, applying Docling and LlamaCloud Services to identify and extract key fields such as invoice number, date, total amount, and vendor details. To ensure the reliability of the extraction process, we establish a robust evaluation framework comprising field-level precision, consistency check failures, and exact match accuracy. The proposed metrics provide a standardized way to compare different extraction methods and highlight strengths and weaknesses in field-specific performance.
\end{abstract}


\section{Introduction}

Invoices are the key document for an organization that stores information on organizational transactions and serves as a proof of purchases and are necessary for accounting and tax purposes \citep{10418145}. Invoices contain information such as purchase date, description of the product , quantity, unit price, and tax deductions. Managing and analyzing invoices is a costly affair for an organization which can be both credit and debit. Extracting and automating information from invoices can significantly increase the productivity of an organization. And hence there is a lot of demand to automate the invoice extraction process. The extracted information must be evaluated on defined metrics to maintain the accuracy of the extraction. With numerous research studies conducted, there are several techniques available for the extraction and evaluation of information extraction from invoices. Although there have been several approaches for the extraction of information from invoices such as graph-based extraction \citep{liu2019graph} \cite{krieger2021information}, grid-based approaches \citep{denk2019bertgrid}, and approaches based on positional embeddings \citep{majumder-etal-2020-representation}, a concrete approach to evaluate the information extracted is still lacking.

In recent years, invoice information extraction (IIE) has gained significant attention as part of the broader field of document information extraction, primarily due to the growing emphasis on business process automation and digital transformation \citep{chen2022xdoc}. Unlike structured documents such as forms, invoices often exhibit heterogeneous layouts depending on vendors, languages, fonts, and formatting practices \citep{borchmann2021due}, making extraction tasks extremely challenging. Rule-driven approaches based on traditional optical character recognition (OCR) were heavily dependent on template-specific heuristics, which had limited generalization \citep{gallo2016deep}. However, machine learning and deep learning methods have introduced more robust techniques for extracting invoice data using multimodal representations of textual, visual, and spatial information \citep{xu2020layoutlm,xu2021layoutxlm}. Despite advances, the challenge of the stent lies in maintaining consistent extraction accuracy and reliability in domains and invoice types. Many state-of-the-art models focus exclusively on improving extraction performance, yet less attention has been paid to establishing standardized evaluation strategies \citep{stanislawek2021kleister}. Given that extracted outputs directly impact downstream tasks such as accounting, fraud detection, and enterprise resource planning (ERP) integration, performance evaluation frameworks are crucial. Standard metrics such as precision, recall, and the F1 score are commonly employed but may not capture the business-critical precision of structured fields such as invoice numbers, tax IDs, or payment terms \citep{trivedi2023secrecy}. Thus, there is an increasing need for domain-specific evaluation methodologies that can assess both syntactic accuracy and semantic consistency in the extraction of invoice information.

Furthermore, large-scale industry datasets for the extraction of invoices remain scarce due to privacy and confidentiality concerns \citep{stanislawek2021kleister}, which limits the benchmarking and reproducibility of research results. Consequently, many works rely on synthetic datasets or proprietary collections, which hinders fair comparison between methods. This research gap highlights the urgency for developing not only effective extraction algorithms, but also robust evaluation frameworks and benchmark datasets. Establishing the EM for invoice data extraction should have a clear ground truth for each targeted field (e.g. invoice number, date, supplier, totals, line-item fields) and measured field level from extraction process \citep{dagdelen2024structured}. A robust EM should cover the accuracy, reliability, business risk, and operational performance from start to finish. Beyond aggregate scores, performance should be stratified by document conditions - digital vs scanned PDF, skew / noise, stamps / handwriting, language, currencies and unseen supplier templates - to measure robustness and generalization, especially for long-tail suppliers where layouts vary widely \citep{krieger2023automated}. Business-centric key performance indicators (KPIs), such as the straight-through processing rate, the manual review rate, and the false auto-approval rate, are critical to reflect the risk of operation and automation effectiveness in real workflows \citep{bardvall2024automating}. Finally, calibration of confidence thresholds with precision recall trade-offs per field, plus continuous benchmarking on fixed canary sets, helps tune auto-approve vs. human review gates and monitor drift over time, ensuring consistent and reproducible comparisons between models and vendors \citep{jijkoun2004information}.

Overall, while numerous extraction approaches continue to evolve, ranging from graph-based to transformer-based representations, the question of performance evaluation under realistic enterprise constraints remains underexplored. This paper seeks to address this research gap by systematically discussing extraction methods, identifying their strengths and weaknesses, and proposing evaluation strategies suitable for the extraction of invoice information. This paper is organized as follows. Section 2 discusses the literature survey. Section 3 discusses the methodologies involved in extracting invoice information. The results and analysis are given in Section 4, after which the paper concludes an outlook on further work in Section 5.

\section{Literature Survey}
Automated invoice information extraction is a critical task in business process automation, aiming to convert unstructured or semi-structured invoice documents into structured data. This process is often called Intelligent Document Processing (IDP), significantly reducing manual efforts, minimizing errors, and accelerating financial workflows. The extraction methods have evolved from traditional template-based methods to sophisticated deep learning models.

\subsection{Traditional and Rule Based Methods}

Early approaches to obtaining invoice information relied heavily on OCR combined with rule-based systems. After converting an invoice image to machine-readable text using an OCR engine like Tesseract, these systems would apply handcrafted rules or templates to locate and extract key information. These methods often used regular expressions (regex) to find specific patterns such as dates (e.g., dd-mm-yyyy) invoice numbers (e.g., INV-), or monetary values. For invoices with a consistent layout, template-based methods are popular. In this approach, zones or coordinates for specific fields (e.g., vendor name, total amount) are predefined for each invoice template. Although effective for a known set of invoice layouts, these systems are brittle and do not generalize to new or unseen invoice formats, which is a significant limitation given the wide diversity of invoice designs \citep{katti2018chargrid}. Maintenance overhead is also substantial, as a new template is required for each new invoice layout.

\subsection{Machine Learning and Deep Learning Approaches}

 Machine learning (ML) and deep learning (DL) approaches are widely used to extract invoice information because of the difficulties of traditional rule-based systems in handling the various and frequently unstructured formats of invoices. The limitations of rule-based systems allow for the adoption of ML and DL models, which can learn to extract information from diverse layouts without explicit rules. By immediately learning patterns from annotated data, ML and DL models can generalize across different formats, in contrast to rule-based techniques that require human development of templates and heuristics for every invoice style. To discover and categorize important invoice fields, machine learning techniques, such as support vector machines (SVMs) and conditional random fields (CRFs), usually rely on designed features. However, by automatically learning rich contextual and geographic representations of invoice content, deep learning techniques, especially those that use transformer-based architectures (e.g., BERT variations) and sequence models such as BiLSTM-CRF, have shown greater results. These models are capable of handling multi-language invoices, noisy OCR outputs, and complex document structures, making them more robust and scalable for real-world applications. As a result, the ML and DL approaches have become the foundation of modern invoice processing systems.

\subsubsection{Text and Layout-Aware Models}

Modern approaches treat invoice extraction as a sequence labeling or question-answering task, incorporating not just the textual content but also the spatial layout of the document. A seminar work by \citep{xu2020layoutlm}, LayoutLM revolutionized document artificial intelligence (AI) by pre-training a transformer-based model on large-scale document image datasets. It extends the BERT architecture by incorporating 2D positional embeddings (bounding-box coordinates) alongside text and visual embeddings. This allows the model to understand the spatial arrangement of words on a page, which is crucial to identifying fields such as headers, line items, and totals. Following LayoutLM, several variants have been proposed. DocFormer \citep{appalaraju2021docformer} introduced a multi-modal self-attention mechanism that learns correlations between text, visual features, and spatial locations more effectively. These models are typically fine-tuned for specific downstream tasks, such as obtaining information from invoices or receipts.

\subsubsection{Graph-Based Methods}

Recognizing that documents possess an inherent graphical, another line of research focused on the application of graph neural network (GNN). In this paradigm, a document is represented as a graph G = (V, E), where the nodes V correspond to text segments (for example, words or text lines) and the edges E represent the spatial and logical relationships between them. Edges are typically constructed on the basis of heuristics such as proximity and vertical/horizontal alignment. A GNN is then applied to this graph structure to learn node representations that encode relational context. The final node embeddings are used for downstream tasks, such as node classification, where each text segment is assigned a semantic label (e.g., invoice date). GNNs are particularly effective at capturing complex, non-local dependencies and modeling tabular structures. The message-passing mechanism inherent to GNNs allows information to propagate across the document, allowing the model to associate entities that are spatially distant but semantically related.

For many businesses, the most critical and challenging data are contained within the line items of an invoice. Graph-based methods are specifically engineered to excel at this. This technique models the invoice as a network of interconnected text elements (a graph). Words are nodes, and their spatial relationships (e.g., being adjacent, being in the same row/column) from the edges. A GNN then analyzes this network to understand complex relationships, such as matching the description of each item of the line to its quantity, unit price, and total \citep{liu2019graph}. The ROI of this method lies in its ability to enable deep process automation and financial control. By accurately extracting line item data, businesses can automate the three-way matching process (comparing the invoice against the purchase order and good receipt note). This is crucial to prevent overpayment, ensure contract compliance, and generate granular spending analytics that can be used for budget optimization and supplier negotiation.

\subsubsection{Large Language Models in Document Information Extraction}

The advent of LLMs has marked a significant paradigm shift in the field of information extraction, moving away from the conventional supervised learning approach that requires extensive task-specific fine-tuning. LLMs leverage vast amounts of pre-trained knowledge to perform complex reasoning and extraction tasks with minimal or no task-specific examples. Traditional DL models, including the layout-aware architectures previously discussed, operate under a supervised learning paradigm. This necessitates the creation of large, domain-specific, and meticulously annotated datasets for fine-tuning, which is both time-consuming and cost-prohibitive.
LLMs introduce the concept of in-context learning \citep{brown2020language}. Instead of updating the model weights, the model is conditioned at inference time using a natural language prompt that describes the task. This approach allows for rapid adaptation to new tasks and document types without a dedicated training phase. The primary modalities for this approach in document extraction are text-based and multi-modal. 
\begin{itemize}

\item Text-based extraction via prompt engineering: The most direct application of LLMs involves processing the text output from an OCR engine. The methodology relies on carefully constructing a prompt that guides the model's behavior.  
\begin{itemize}
\item Zero-shot learning: In this setting, the LLM is given a high-level task description without any concrete examples. The model relies entirely on its pre-trained knowledge to understand the task and extract the required information.  
\item Few-shot learning: To improve performance and guide the model towards the desired output format, the prompt is augmented with a few examples (typically 1 to 5) of input documents and their corresponding structured outputs. This technique has been shown to significantly enhance extraction accuracy (Brown et al., 2020).
\end{itemize}
Prompt engineering has emerged as a critical discipline in itself. Structure, wording, and examples within the prompt can have a substantial impact on the quality of the extraction. A typical prompt for invoice extraction would include the role, the task instructions, the desired output format (e.g., JSON schema), the few-shot examples (if any), and the OCR-extracted text of the target invoice.

\item Multimodal models for visual document understanding; a limitation of the text-based approach is its dependency on the quality of the upstream OCR engine. Errors or structural misinterpretations from the OCR phase are propagated to the LLM. Multimodal LLMs, also known as Vision-Language Models (VLMs), mitigate this by processing the document image directly. These models, such as GPT-4V \citep{openai20234v} and open-source alternatives such as LLaVA \citep{liu2023visual}, integrate a vision encoder with a large language model. The vision encoder processes the raw image to extract visual features, which are then passed to the language model along with the textual prompt. This allows the model to ground the text in its visual context, using layout, font styles, logos, and table structures that are often lost in plain text. This approach is more closely aligned with the comprehension of human-like documents.

\end{itemize}

\section{Methodology}

The study implemented in this paper employs a comparative experimental design to evaluate the performance of two distinct methodological paradigms of obtaining information from invoices: a dedicated document understanding model (Dockling) and a LlamaExtractor. The methodology adopted in the work tragets key value fields and line items typically present in business invoices, aligned with community benchmarks on key information localization and extraction (KILE) and line item recognition (LIR) from DocILE and related tasks. The following subsections detail the implementation specifics and evaluation metrics for each approach.

\subsection{Dataset and Preprocessing}

Effective data set preparation is foundational for the performance and generalization of invoice information extraction systems. To address the scarcity of annotated resources and ensure model robustness, prior work emphasizes the importance of creating diverse, well-annotated invoice datasets. A curated data set of 102 digitally born and scanned invoice documents in PDF and JPEG format is utilized. Invoices vary in layout, language (e.g., English, German) and complexity. Each invoice is manually annotated to create a ground truth JSON schema containing key fields such as invoice\_number, issue\_date, due\_date, seller\_name, seller\_address, total\_amount, tax\_amount, and line\_items. Curation part also invoved the variability of the input image by rotating the images at different angles to creat few synthetic dataset \citep{pv2025generating}. Comparsion of ground truth with the results obtained from different methods listed helps us in setting up the evaluation metrics. This combined approach—diverse real data, synthetic augmentation, and robust preprocessing lays a strong foundation for a reliable evaluation of extraction methods across varied invoice instances.

\subsection{Method 1: Docling-Based Extraction}

\begin{figure}[h!]
    \centering
    \includegraphics[width=\textwidth]{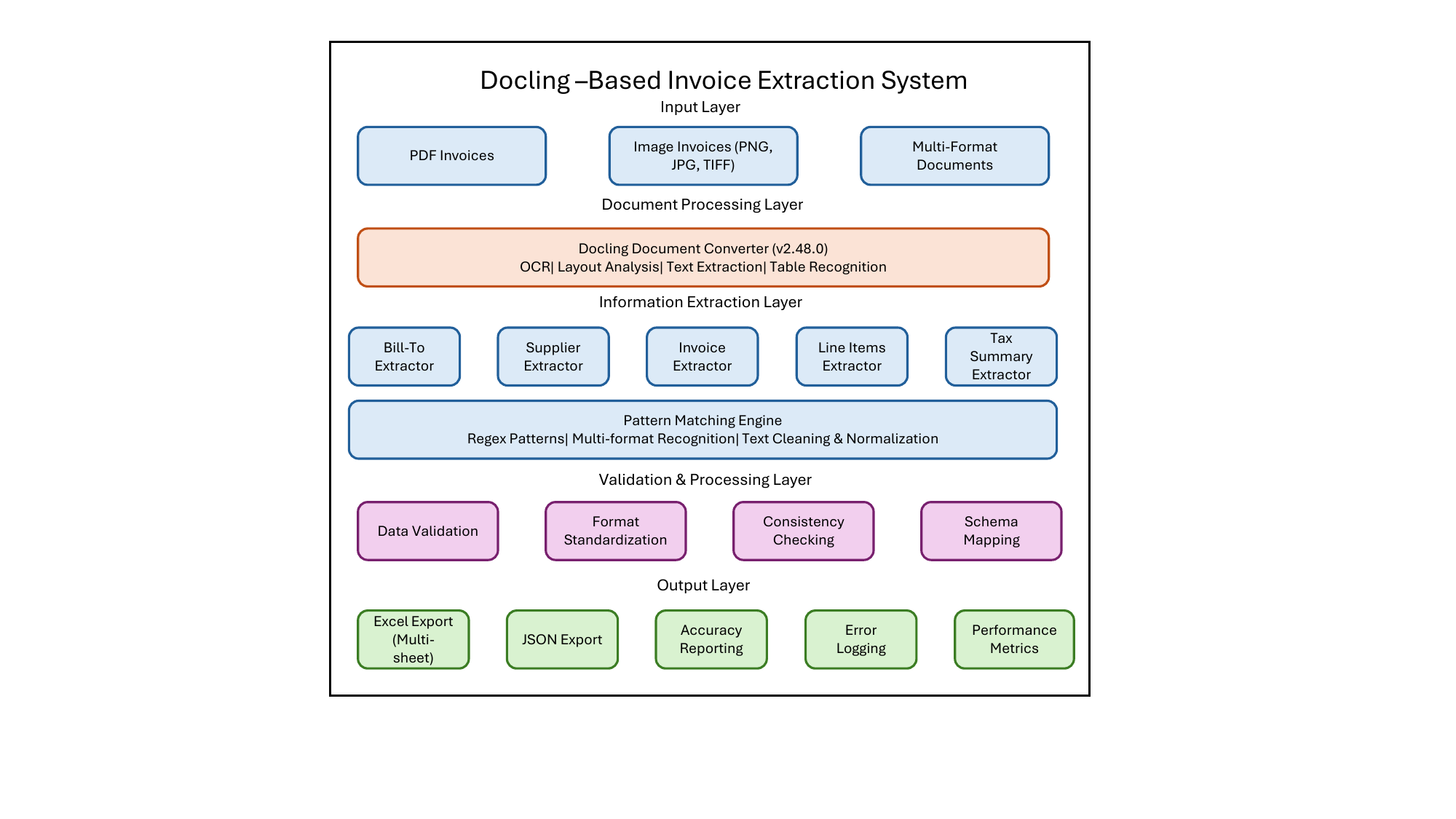}
    \caption{Architecture Diagram of the  Docling System}
    \label{fig:docling_architecture}
\end{figure}

Docling is used to parse invoices from diverse formats (PDF, images, Office files) into a unified Docling document representation and export machine-readable JSON/Markdown suitable for downstream extraction. It's an end-to-end document understanding model specifically designed to parse complex documents into structured JSON data. Its architecture bypasses traditional OCR and layout detection pipelines, treating the document as a unified modality. We utilized a pre-trained Docling-base model. Each document is parsed into blocks (text, tables, images) with reading order and structure preserved, then exported to JSON to standardize inputs for rule-based mapping. Table regions are identified and normalized to rows/columns to support line-item extraction downstream. Docling’s CV-first parsing reduces OCR error propagation in these steps. Key fields (e.g., vendor\_name, dates, totals) are localized using layout-aware text blocks and table structures; extracted values are serialized with bounding boxes to satisfy the KILE/LIR evaluation The model's strength lies in its ability to understand spatial relationships and textual semantics concurrently \citep{appalaraju2021docformer}. The model is prompted with a textual description of the desired output schema. The architectural diagram representing the overall extraction flow is shown in Fig.~\ref{fig:docling_architecture}

\subsection{Method 2: Llama Extractor}

\begin{figure}[h!]
    \centering
    \includegraphics[width=\textwidth]{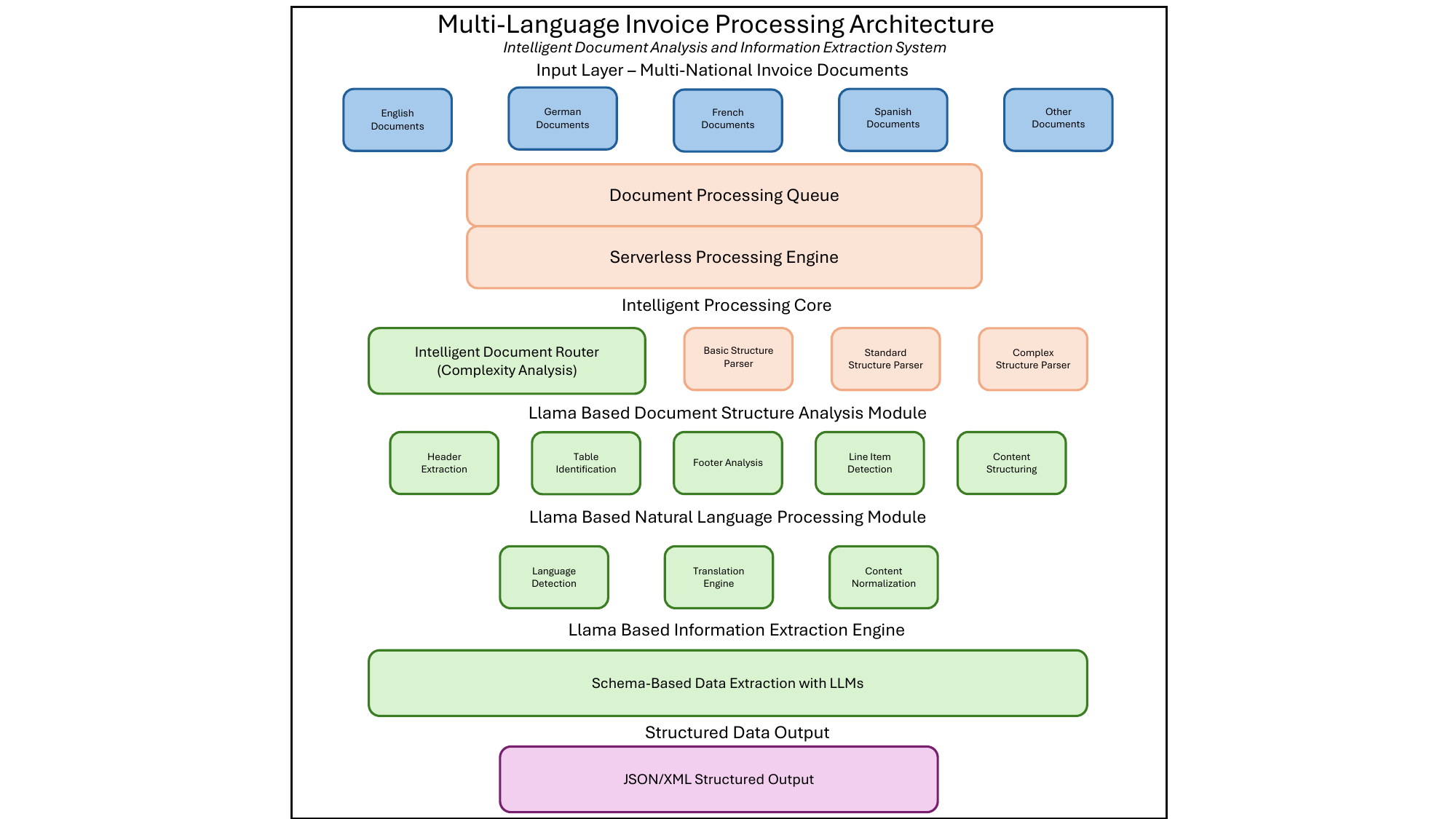}
    \caption{Architecture Diagram of LlamaExtractor}
    \label{fig:llama_architecture}
\end{figure}

The LlamaExtractor uses schema-guided, LLM-based structured extraction to return deterministic JSON that conforms to a declared schema for invoice fields and line items.
To enforce structure, the extractor relies on structured output capabilities, such as JSON Schema-constrained decoding and function/tool signatures, that guarantee type-safe outputs and minimize parsing failures. Implementations are performed with LlamaIndex structured extraction utilities or managed extraction services (e.g., LlamaExtract) that take a Pydantic/JSON schema and emit well-typed JSON from unstructured or semi-structured inputs. Schema-constrained output is achieved by attaching the schema to the LLM (e.g., through LlamaIndex structured outputs or the Llama API’s structured output interface), which steers decoding to produce only valid fields, types, and nesting. A predefined 'invoice' schema is commonly available in Llama-driven document agents and can be used as is or extended to custom fields without additional fine-tuning. Prompts describe the task, the schema, and any field-level instructions (e.g., normalization hints, required formats, and units), optionally augmented by few-shot exemplars that illustrate desired JSON outputs for invoices with and without line items. The LLM receives the parsed invoice text as context and is instructed to extract only the requested fields and to leave any missing fields null or empty as defined by the schema. The extractor can produce a confidence score and brief rationale per field as part of the structured output schema to facilitate triage, auditing, and human-in-the-loop workflows. The architectural diagram representing the general extraction flow is shown in Fig.~\ref{fig:llama_architecture}. The workflow of extracting information is as follows - 

\begin{itemize}
    \item Invoices from different countries with different languages enter the system.
    \item A custom router in place routes the invoice to different parsing/extraction strategies depending on the complexity of the invoice.
    \item After routing the invoice to a particular channel, parsing prcocess of the invoice start where it identifies the documents key regions (headers, tables, footers, line items) and prepares the structured context input to for extraction.
    \item After parsing, the system detects the invoice language and triggers translation workflows if the content is not in English.
    \item Once the translation of the invoice text into English is completed, the structured data extraction process is initiated. This process relies on a predefined schema which outlines the expected fields such as invoice number, date, vendor details, and line items.
\end{itemize}

\subsection{Evaluation Metrics}

Performance is evaluated at field and document levels, with metrics aligned with DocILE tasks: key information localization / extraction and line item recognition, including zero / few shot analyzes \citep{vsimsa2023overview}. 

 \begin{itemize}
\item Field-level metrics: These metrics assess the accuracy of identifying and extracting structured header fields such as invoice numbers, dates, buyer/seller details, and totals. Evaluation includes:

\begin{itemize}
    \item Exact match and relaxed match scoring to account for strict correctness versus partial or fuzzy matches (e.g., text normalization or OCR-induced variants).
    \item Tolerance windows for numeric fields, such as amounts and dates, allow for acceptable deviations due to formatting or rounding.
    \item F1 scores for binary classification of field presence and correctness, enabling nuanced evaluation of whether fields are correctly detected, even if their values are partially incorrect.
\end{itemize}
 
\item Line item metrics: Given the complexity of tabular data in financial documents, line-item performance is evaluated through:

\begin{itemize}
    \item Row-wise assignment accuracy, which measures how well the model groups and extracts values into coherent line items.
    \item Tolerance-based numerical checks for quantities and monetary values, taking into account common OCR or format-related errors.
    \item Table-level completeness, which evaluates the extent to which the model captures all relevant line-items without omissions or duplication.
\end{itemize}
 
\item Robustness: Beyond static accuracy metrics, robustness is a critical consideration for real-world deployments. Evaluations include:

\begin{itemize}
    \item Cross-vendor performance, measuring consistency when documents originate from multiple sources or layouts.
    \item Template generalization in zero-shot (unseen layouts) and few-shot (limited training examples) settings, to test adaptability across diverse document types and vendors.
\end{itemize}
 \end{itemize}

\section{Results and Analysis}

This section presents a comparative evaluation of two distinct methods applied to the task of obtaining invoice information: Docling and LlamaExtractor. The performance of each approach is assessed based on accuracy, efficiency, and robustness in extracting key invoice components such as invoice number, date, vendor details, and total amount. The results have been anlyzed on two different parameters listed below - 

\begin{itemize}
    \item Accuracy distribution measures the percentage of invoice fields correctly extracted by each method, highlighting strengths and weaknesses across different data points.
    \item Consistency Check Failures count instances where extracted data violates predefined business rules, such as mismatched gross and net amounts, indicating extraction errors.
\end{itemize}

The extraction and evaluation schema consists of several key entities, each representing different aspects of the invoice data. The bill-to-entity includes fields related to the customer or business unit responsible for payment. The supplier entity captures vendor-specific information such as tax IDs and bank details. The invoice entity encompasses general transactional details, including invoice numbers, dates, and payment terms. Detailed transaction lines are covered under the Line Item entity, which records individual services or products with pricing and tax information. Finally, the tax line summary entity summarizes the tax-related fields in the invoice to ensure compliance and accurate reporting.

\subsection{Result - Docling}

The documenting methodology is evaluated on a benchmark invoice data set, focusing on field-level accuracy and document-level consistency along with consistency checks.

\begin{itemize}
\item The methodology achieved an overall accuracy of 63\%, demonstrating moderate performance in recognizing and extracting key information from the invoice.
\item Accuracy varied significantly across entities, with entities like invoice\_name, invoice\_number showing the highest extraction accuracy at 90\%, benefiting from the Docling's layout understanding and parsing approach.
\item The tax\_amount, net\_amount, currency\_format entities exhibited lower accuracy at 60\%, primarily due to limitations in handling unstructured text fields, complex table structures, multi-line descriptions.
\item The Line Item entity, being the most granular and diverse in content, achieved an accuracy of 58\%. This performance reflects the difficulty of documenting with complex nested table layouts.
\end{itemize}
Beyond individual field accuracy, consistency checks were performed to ensure that the extracted data adhered to the logical and business rules encoded in the schema. For example, net amount + tax amount + roundoff amount = gross amount in the invoice entity. The Docling methodology achieved a 80\% pass rate on these consistency checks, with 20\% of extractions failing due to incorrect numerical field extraction leading to calculation mismatches/misalignment between line item totals and invoice summaries/errors in tax calculation fields. Further analysis revealed that most consistency failures were concentrated in invoices with complex multi-tax scenarios and failures predominantly occurred in invoices with non-standard formatting.

\subsection{Result - LlamaExtractor}

The LlamaExtractor is evaluated on a curated dataset annotated with the predefined schema covering five key entities: bill-to, supplier, invoice, line item, and tax line summary. The overall accuracy of the LlamaExtractor is calculated as the proportion of correctly extracted fields relative to the total annotated fields throughout the dataset.

\begin{itemize}
    \item The extractor achieved an accuracy of 94\%, demonstrating strong performance in recognizing and extracting key invoice information.

 \item Accuracy varied between entities, with the invoice and bill to entities showing the highest extraction precision due to their relatively standardized field formats.

\item The line item entity, being the most granular and diverse in content, exhibited a slightly lower accuracy of 91\%, reflecting challenges in handling variable descriptions and numeric values on different invoice layouts.
\end{itemize}

Beyond individual field accuracy, consistency checks were performed to ensure that the extracted data adhered to the logical and business rules encoded in the schema. For example, net amount + tax amount + round-off amount = gross amount in the invoice entity.

The LlamaExtractor passed 93\% of these general consistency checks.

\subsection{Comparative Analysis}

The comparative evaluation between the Docling and the LlamaExtractor method highlights distinct trade-offs in accuracy, efficiency, technical architecture, and practical usability for automated invoice data extraction.

\subsubsection{Overall Performance}

The LlamaExtractor shows superior accuracy, achieving an overall extraction accuracy of 94\%, compared to 63\% for the Docling method. This significant performance gap underscores the stronger ability of the LlamaParser model to interpret and structure invoice content in various formats. 

In terms of processing efficiency, the Docling method exhibits faster average processing times (10 seconds per page) relative to the LlamaExtractor (30 seconds per page). This suggests that Docling prioritizes lightweight processing, likely due to its simpler OCR and layout-based approach, while the LlamaExtractor’s deep language modeling introduces higher computational overhead but delivers better semantic understanding.

Table~\ref{tab:overall_consistency_comparison} shows the overall comparitive performance between LlamaExtractor and Docling method.

\begin{table}[htbp]
\centering
\caption{Overall Performance and Consistency Validation Comparison}
\label{tab:overall_consistency_comparison}
\small
\resizebox{\textwidth}{!}{%
\begin{tabular}{|l|c|c|c|c|}
\hline
\textbf{Method} & \textbf{Overall Accuracy (\%)} & \textbf{Processing Time (avg/page)} & \textbf{Consistency Check Pass Rate (\%)} & \textbf{Mathematical Validation Errors(\%)} \\
\hline
Llama Extractor & 94 & 30 sec & 93 & 5 \\
\hline
Docling Method & 63 & 10 sec & 80 & 20 \\
\hline
\end{tabular}%
}
\end{table}

\subsubsection{Consistency Validation}

The LlamaExtractor achieves a consistency check pass rate 93\%, indicating highly stable and reliable output, even across invoices with varying layouts or linguistic nuances. In contrast, the Docling method records an 80\% pass rate, revealing occasional inconsistencies, especially in field alignment and numerical validation. 

Furthermore, the LlamaExtractor recorded 5\% mathematical validation errors, while the Docling exhibited 20\% errors, suggesting that the internal reasoning capabilities of the Llama enhance the logical verification of the extracted numeric data, such as totals, taxes and line item calculations.

Table~\ref{tab:overall_consistency_comparison} shows the comparitive consistency validation between LlamaExtractor and Docling method.

\subsubsection{Technical Characteristics}

Both methods support similar input formats (raw text, PDF, JPG), enabling flexible document ingestion. However, their underlying architectures differ significantly. 

The LlamaExtractor operates as an LLM, using contextual understanding and reasoning to interpret complex invoice structures. This allows it to be generalized across unseen templates without explicit layout training. 

In contrast, the documenting method depends on layout analysis and OCR pipelines, making it more deterministic, but less adaptable to diverse invoice templates. Consequently, Llama requires only minimal preprocessing, while Docling necessitates explicit document structuring and OCR preprocessing to function optimally. 

From a computational point of view, the LlamaExtractor requires medium resources, while the low resource consumption of Docling makes it more suitable for small- or large-scale deployments with limited computational infrastructure.

Table~\ref{tab:technical_characteristics} shows the comparison between technical characteristics of LlamaExtractor and Docling method.

\begin{table}[htbp]
\centering
\caption{Technical Characteristics of Invoice Extraction Methods}
\label{tab:technical_characteristics}
\small
\resizebox{\textwidth}{!}{%
\begin{tabular}{|l|c|c|c|c|}
\hline
\textbf{Method} & \textbf{Input Format} & \textbf{Model Architecture} & \textbf{Pre-processing Required} & \textbf{Computational Resources} \\
\hline
Llama Extractor & Raw Text/PDF/JPG & Large Language Model & Minimal & Medium \\
\hline
Docling Method & Raw Text/PDF/JPG & Layout Analysis + OCR & Required & Low \\
\hline
\end{tabular}%
}
\end{table}

\subsubsection{Error Analysis}

The error patterns further reveal the fundamental differences in the two systems. The primary sources of error of the LlamaExtractor stem from variable field descriptions and numeric formatting variations, such as unconventional representations of dates or currency. 

On the other hand, docing errors arise predominantly from mathematical validation issues and incomplete line item extractions, which can be attributed to OCR misreads or inconsistent mapping of the bounding box. 

In terms of structural robustness, the LlamaExtractor handles complex layouts more effectively (rated “Good”) compared to Docling (“Fair”). Additionally, Llama supports multi-page document parsing, enabling end-to-end extraction from lengthy invoices, while Docling’s support is limited, reducing its applicability for multi-page commercial invoices.

Table~\ref{tab:error_analysis} shows the comparitive error analysis of LlamaExtractor and Docling method.

\begin{table}[htbp]
\centering
\caption{Error Analysis of Llama Extractor and Docling Method}
\label{tab:error_analysis}
\small
\resizebox{\textwidth}{!}{%
\begin{tabular}{|l|c|c|c|}
\hline
\textbf{Method} & \textbf{Primary Error Sources} & \textbf{Complex Layout Handling} & \textbf{Multi-page Document Support} \\
\hline
Llama Extractor & Variable descriptions, numeric formatting & Good & Yes \\
\hline
Docling Method & Mathematical validation, line item description & Fair & Limited \\
\hline
\end{tabular}%
}
\end{table}

\subsubsection{Practical Considerations}

From an operational standpoint, the LlamaExtractor exhibits medium setup complexity, primarily due to model integration and configuration requirements. The Docling method, being a rule-driven OCR pipeline, offers low setup complexity and can be more easily deployed in traditional environments. 

Regarding scalability, the LlamaExtractor demonstrates excellent scalability, benefiting from its ability to generalize without layout retraining. Docling, though functional, scales only fairly, as its OCR-based pipeline may require custom adjustments for new formats. 

In terms of cost efficiency, Docling performs better (High) because of its lower resource demands and simpler architecture, whereas the LlamaExtractor incurs medium operational costs due to higher inference time and model complexity. Both systems, however, are capable of real-time processing, making them viable for near-instantaneous invoice validation and automation workflows.

Table~\ref{tab:practical_considerations} shows the practical considerations for LlamaExtractor and Docling method.

\begin{table}[htbp]
\centering
\caption{Practical Considerations for Deployment and Scalability}
\label{tab:practical_considerations}
\small
\resizebox{\textwidth}{!}{%
\begin{tabular}{|l|c|c|c|c|}
\hline
\textbf{Method} & \textbf{Setup Complexity} & \textbf{Scalability} & \textbf{Cost Efficiency} & \textbf{Real-time Processing} \\
\hline
Llama Extractor & Medium & Excellent & Medium & Yes \\
\hline
Docling Method & Low & Fair & High & Yes \\
\hline
\end{tabular}%
}
\end{table}

\subsubsection{Summary}

In summary, the LlamaExtractor provides a more intelligent, semantically aware, and accurate solution for invoice data extraction, particularly suited for enterprises requiring high reliability and scalability across varied invoice templates. The Docling method, while less accurate, offers a lightweight, faster, and more cost-efficient alternative, suitable for applications where speed and simplicity outweigh the need for high precision. 

Thus, the choice between the two depends on the specific operational priorities: accuracy and adaptability (Llama) versus cost-effectiveness (Docling).

\section{Conclusion and Future Discussion}

This study presented a comprehensive comparative analysis between two different invoice extraction methodologies: the LlamaExtractor, an LLM-based system, and the Docling method, an OCR-based approach and layout-driven approach. Through detailed evaluation across performance, consistency, technical characteristics, and practical dimensions, the analysis highlights that the LlamaExtractor consistently outperforms traditional document parsing frameworks in terms of accuracy, semantic understanding, and robustness across diverse invoice structures.

The results show that while the LlamaExtractor achieves a significantly higher overall accuracy of 94\%, it does so with moderately higher computational requirements and processing time. Its reasoning-driven architecture ensures near-perfect consistency validation and eliminates most numerical or structural errors, making it particularly suitable for enterprise-grade invoice automation systems where precision and reliability are critical. In contrast, the Docling Method, though less accurate, offers faster execution and lower computational cost, making it better suited for lightweight or cost-sensitive deployments.

In general, this comparative assessment underscores the trade-off between semantic accuracy and computational efficiency. It establishes that modern LLM-based extractors, when properly optimized, can redefine the state of automated document understanding, bridging the gap between human-like reasoning and structured data extraction from unstructured financial documents.

Future research can extend this work in several promising directions. First, integrating a hybrid extraction pipeline that combines the semantic reasoning of LLMs with the structured layout awareness of OCR systems could yield a more balanced and efficient approach. Such hybrid models may significantly reduce latency while maintaining high accuracy.

Second, exploring domain-adaptive fine-tuning of language models in invoice-specific corpora could enhance contextual accuracy, particularly for vendor-specific templates, regional tax formats, and multilingual invoices. Incorporating self-correction mechanisms or confidence-based validation layers could further improve reliability during large-scale deployments.

Another potential avenue involves the development of a benchmark data set for the evaluation of invoice extraction that includes various layouts, languages, and industries. This would enable standardized comparison across methods and foster reproducibility in research.

Lastly, expanding this investigation to include other financial documents such as purchase orders, receipts, and contracts could provide a broader understanding of how LLM-based and OCR-based systems generalize across related domains. Such explorations would be instrumental in evolving towards fully autonomous document understanding systems within enterprise ecosystems.

In summary, while LLM-based extraction methods, such as the LlamaExtractor, have shown clear advantages, future work should focus on optimizing their performance, scalability, and cost-efficiency to ensure widespread adoption in real-world production-grade invoice processing pipelines.

\bibliographystyle{unsrtnat}







\end{document}